\documentclass[conference]{IEEEtran}
\usepackage{times}
\usepackage[numbers]{natbib}
\usepackage{multicol}
\usepackage[bookmarks=true]{hyperref}
\hypersetup{
    colorlinks=true,
    linkcolor=blue,
    filecolor=magenta,      
    urlcolor=cyan
    }
\usepackage{graphics} 
\usepackage{epsfig} 
\usepackage{mathptmx} 
\usepackage{amsmath} 
\usepackage{amssymb}  
\usepackage{siunitx} 
\usepackage{textcomp}
\usepackage{gensymb} 
\usepackage{booktabs}
\usepackage{multirow}
\usepackage{xcolor}
\usepackage{algorithm}
\usepackage{algpseudocode}
\let\labelindent\relax
\usepackage{enumitem}
\usepackage{xspace}
\usepackage{float}
\usepackage{comment}
\usepackage{mathtools}
\usepackage{soul}
\usepackage{natbib}
\usepackage[most]{tcolorbox}
\usepackage{xcolor}

\tcbset{
    greenbox/.style={
        colback=green!20, 
        colframe=green!60!black, 
        coltext=green!60!black, 
        boxrule=0.5mm,
        arc=2mm, 
        auto outer arc,
        boxsep=5pt,
        left=2mm,
        right=2mm,
        top=2mm,
        bottom=2mm,
    }
}

\usepackage{makecell}
\usepackage{rotating}
\usepackage[percent]{overpic}
\usepackage{contour}
\usepackage{courier}

\usepackage{subcaption} 
\usepackage[font=small]{caption}
\usepackage[percent]{overpic}

\usepackage[11pt]{moresize}

\usepackage{pifont}
\definecolor{MyDarkBlue}{rgb}{0,0.08,1}
\definecolor{airforceblue}{rgb}{0.36, 0.54, 0.66}
\definecolor{MyDarkGreen}{rgb}{0.02,0.6,0.02}
\definecolor{MyDarkRed}{rgb}{0.8,0.02,0.02}
\definecolor{MyDarkOrange}{rgb}{0.40,0.2,0.02}
\definecolor{MyPurple}{RGB}{111,0,255}
\definecolor{MyRed}{rgb}{1.0,0.0,0.0}
\definecolor{MyGold}{rgb}{0.75,0.6,0.12}
\definecolor{MyDarkgray}{rgb}{0.66, 0.66, 0.66}
\definecolor{MyPink}{rgb}{0.9, 0.33, 0.5}
\definecolor{MyCyan}{rgb}{0., 0.4, 0.4}

\newcommand{\ours}{HoMMI\xspace}

\definecolor{guidance_green}{RGB}{12,131,27}
\definecolor{theme_orange}{RGB}{255,138,0}
\definecolor{theme_blue}{RGB}{67,99,216}
\definecolor{theme_taro}{RGB}{219,176,234}

\definecolor{pure_green}{RGB}{0,255,0}
\definecolor{pure_red}{RGB}{255,0,0}

\makeatletter
\DeclareRobustCommand\onedot{\futurelet\@let@token\@onedot}
\def\@onedot{\ifx\@let@token.\else.\null\fi\xspace}

\makeatother

\usepackage[most]{tcolorbox}

\usepackage{xparse}
\usepackage{lipsum}

\def\exampletext{Example} 

\NewDocumentEnvironment{testexample}{ O{} }
{
    \colorlet{colexam}{theme_blue} 
    \newtcolorbox[use counter=testexample]{testexamplebox}{%
        empty,
        title={\exampletext: #1},
        attach boxed title to top left,
        minipage boxed title,
        boxed title style={empty,size=minimal,toprule=0pt,top=2pt,left=2mm,overlay={}},
        coltitle=colexam,fonttitle=\bfseries,
        before=\par\nonskip\noindent,parbox=false,boxsep=0pt,left=2mm,right=0mm,top=2pt,breakable,pad at break=0mm,
        before upper=\csname @totalleftmargin\endcsname0pt, 
        after=\par\nonskip,
        overlay unbroken={\draw[colexam,line width=.5pt] ([xshift=-0pt]title.north west) -- ([xshift=-0pt]frame.south west); },
        overlay first={\draw[colexam,line width=.5pt] ([xshift=-0pt]title.north west) -- ([xshift=-0pt]frame.south west); },
        overlay middle={\draw[colexam,line width=.5pt] ([xshift=-0pt]frame.north west) -- ([xshift=-0pt]frame.south west); },
        overlay last={\draw[colexam,line width=.5pt] ([xshift=-0pt]frame.north west) -- ([xshift=-0pt]frame.south west); },%
    }
    \begin{testexamplebox}}
        {\end{testexamplebox}\endlist}

\AfterEndEnvironment{testexample}{\color{black}}

\DeclareMathAlphabet{\mathcal}{OMS}{cmsy}{m}{n}

\usepackage{wrapfig}
\BeforeBeginEnvironment{wrapfigure}{\setlength{\intextsep}{0pt}}
\AfterEndEnvironment{wrapfigure}{\setlength{\intextsep}{0pt}}

\begin{document}

\title{HoMMI: Learning Whole-Body Mobile Manipulation from Human Demonstrations}

\author{
Xiaomeng Xu$^{1,2}$\quad
Jisang Park$^{1}$\quad
Han Zhang$^{1}$\quad
Eric Cousineau$^{2}$\quad
Aditya Bhat$^{2}$\quad
Jose Barreiros$^{2}$\quad \\
Dian Wang$^{1}$\quad
Jeannette Bohg$^{1}$\quad
Shuran Song$^{1}$\quad \\\\
$^{1}$Stanford University\quad
$^{2}$Toyota Research Institute \\
\large{\href{https://hommi-robot.github.io}{https://hommi-robot.github.io}}
}



%

\twocolumn[{
\renewcommand\twocolumn[1][]{#1}
\maketitle
\includegraphics[width=\linewidth]{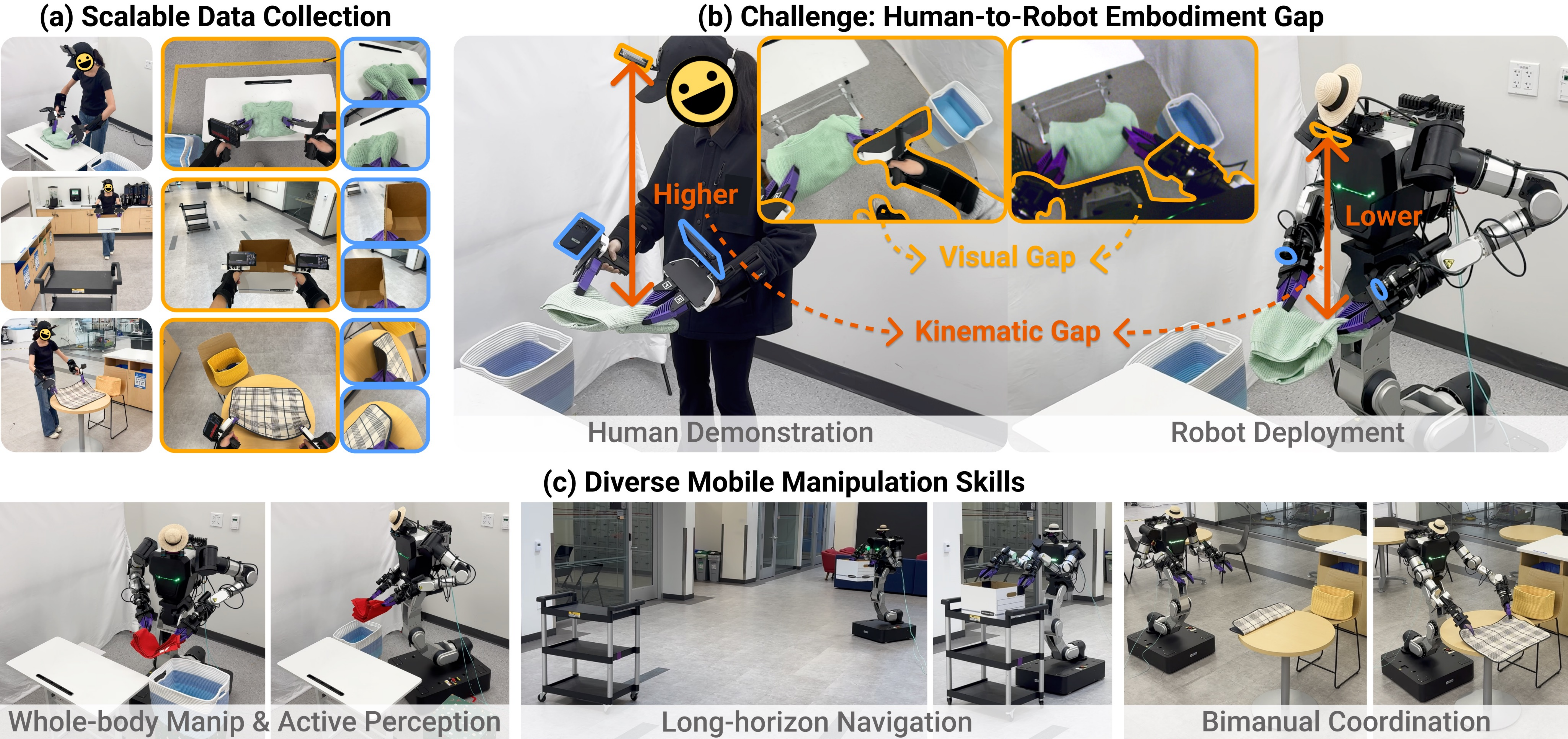}
\captionof{figure}{
\textbf{W\underline{ho}le-Body \underline{M}obile \underline{M}anipulation \underline{I}nterface (\ours).}
(a) We extend UMI with egocentric sensing to enable scalable \textit{\textbf{mobile}} manipulation with \textit{\textbf{active perception}} -- capabilities that cannot be achieved with the original UMI.
(b) However, the new egocentric view creates a substantial embodiment gap in both observation and action space, making policy transfer difficult. 
(c) We bridge this embodiment gap by carefully redesigning the visual and action representations and integrating them with a constraint-aware whole-body controller.
Together, \ours is able to learn diverse mobile manipulation skills directly from human demonstrations, without \textbf{\textit{any}} robot teleoperation data.
\label{fig:teaser}}

\vspace{2mm}
}]

\begin{abstract}
We present Whole-Body Mobile Manipulation Interface (HoMMI), a data collection and policy learning framework that learns whole-body mobile manipulation directly from robot-free human demonstrations. We augment UMI interfaces with egocentric sensing to capture the global context required for mobile manipulation, enabling portable, robot-free, and scalable data collection. However, naively incorporating egocentric sensing introduces a larger human-to-robot embodiment gap in both observation and action spaces, making policy transfer difficult.  We explicitly bridge this gap with a cross-embodiment hand-eye policy design, including an embodiment agnostic visual representation; a relaxed head action representation; and a whole-body controller that realizes hand-eye trajectories through coordinated whole-body motion under robot-specific physical constraints. Together, these enable long-horizon mobile manipulation tasks requiring bimanual and whole-body coordination, navigation, and active perception. All code, data, and hardware design are publicly available at \href{https://github.com/xxm19/hommi}{https://github.com/xxm19/hommi}. 
\end{abstract}

\IEEEpeerreviewmaketitle

\section{Introduction} 
Achieving generalizable and effective mobile manipulation requires seamless \textbf{whole-body coordination}, which consists of coordinating diverse \emph{sensory} inputs (e.g., egocentric head-mounted cameras to eye-in-hand wrist cameras) and complex \emph{action} spaces (e.g., between the arms, torso, head, and base movements). 
Manually programming such intricate coordination for the vast variety of real-world tasks is prohibitively difficult, making learning from human a promising alternative.

However, existing human demonstration paradigms mostly rely on robot teleoperation, which is expensive, slow, and unintuitive to deploy for mobile manipulators across diverse real-world settings. 
Handheld data collection devices such as UMI~\cite{chi2024universal} offer a more scalable solution. They essentially learn end-effector motions through handheld grippers with wrist-mounted camera observations, allowing portable and robot-free demonstration collection. However, wrist-centric sensing provides only local views around the end-effectors and often under-observes the global context needed for navigation, bimanual coordination, and task progress tracking. 

Adding an egocentric view (i.e., head-mounted camera) is a natural solution to fill this gap. By capturing the broader workspace, the spatial relationship between hands, as well as humans' active perception behaviors, egocentric views provide critical information that wrist cameras lack. However, \textit{\textbf{naively incorporating egocentric sensing into UMI framework introduces a larger human-to-robot embodiment gap}}, including:

\begin{itemize}[leftmargin=4mm]
    \item \emph{Visual gap:} Human and robot arms differ in appearance, and egocentric viewpoints vary due to height discrepancies between human and robot embodiments.
    \item \emph{Kinematic gap:} Humans and robots differ in body morphology and neck degrees of freedom.
    Directly regressing and tracking both hands and head 6-DoF trajectories often yield infeasible robot motions.
\end{itemize}

As a result, prior egocentric systems either rely on additional teleoperation data for action grounding~\cite{kareer2025egomimic, zhu2026emma}, or restrict the application domain to fixed-base bimanual manipulation without whole-body coordination~\cite{zeng2025activeumi, yu2025egomi}.
This paper aims to \textit{\textbf{scale mobile manipulation learning by augmenting the UMI framework with egocentric observation, while explicitly bridging the embodiment gap.}} Our system highlights the following key technical contributions:

\begin{itemize}[leftmargin=4mm]
    \item \textbf{\ours Data Collection System:}  We extend the bimanual UMI framework with a head-mounted camera. By integrating the iPhone ARKit, the system enables synchronous capture of multi-view video and 6-DoF poses within a unified and globally consistent coordinate frame.
    
    \item \textbf{Embodiment-Agnostic Vision Representations:} To bridge the observation gap, we use a 3D visual representation for egocentric observations. This allows us to use embodiment-agnostic coordinate frames (i.e., end-effector frame), and remove embodiment-specific observations (e.g., demonstrator's arms and body), mitigating appearance and viewpoint mismatches.
    
    \item \textbf{Relaxed Head Action Representation:} 
    Since our egocentric representation is view-agnostic, we represent the robot gaze as a ``3D look-at point'' to bridge the kinematic gap. Compared with directly copying the 6-DoF head poses from humans, which is often kinematically incompatible with robot hardware, this relaxed action representation enables \emph{effective} transfer of active perception strategies to robots with disparate heights and joint constraints, without sacrificing the tracking accuracy of end-effectors.

    \item \textbf{Constraint-Aware Whole-Body Control:} 
    We design a whole-body controller that can coordinate whole-body motions to \emph{precisely} track end-effector trajectories for accurate manipulation, while respecting the unique constraints in a bimanual mobile robot system for stable and safe motions. 
\end{itemize}

Together, these ideas enable a scalable, in-the-wild human demonstration collection that is directly transferable to real robots. We demonstrate that our system achieves precise, long-horizon, and spatially complex whole-body mobile manipulation tasks, including active search, manipulation, and navigation across large workspaces.
\begin{figure*}[t]
    \centering
    \includegraphics[width=\linewidth]{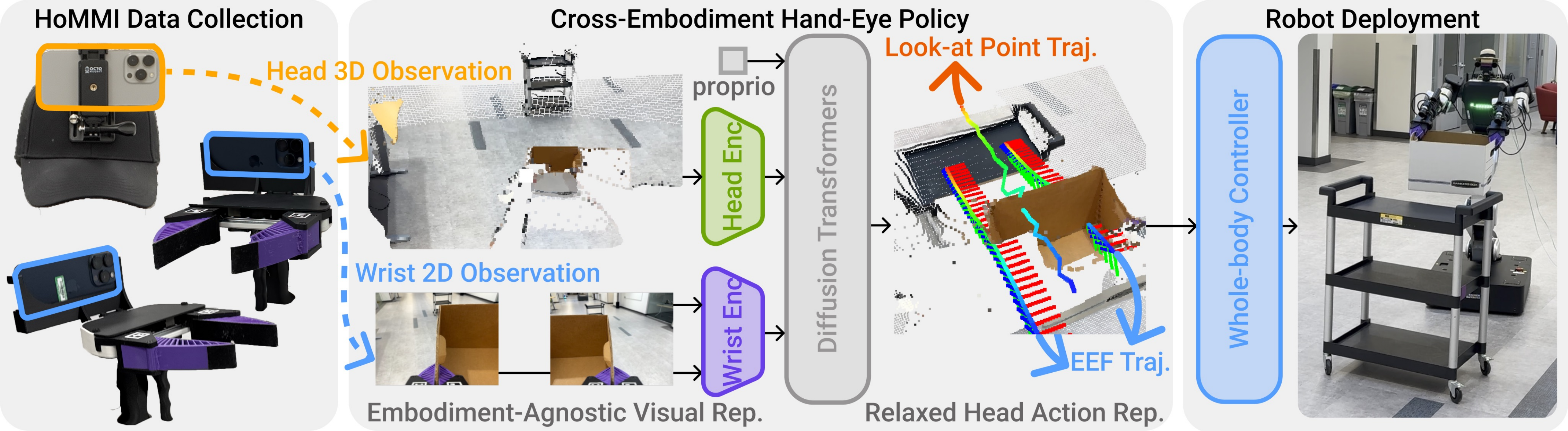}
    \caption{\textbf{\ours System Overview.} We learn whole-body mobile manipulation from human demonstrations with an intuitive data collection interface (\S~\ref{sec:data_collection}), a cross-embodiment policy design with an embodiment-agnostic visual representation and a relaxed head action representation (\S~\ref{sec:policy}), and a whole-body controller that achieves hand-eye tracking through whole-body motions respecting physical constraints (\S~\ref{sec:wbc}).}
    \label{fig:pipeline}
    \vspace{-6mm}
\end{figure*}

\section{Related Work}
\vspace{-1mm}
\subsection{Data Collection Interfaces for Robot Learning}
\vspace{-1mm}
Robot learning from demonstrations traditionally relies on teleoperation~\cite{wu2024gello, XuX-RSS-25, xu2025compliant, xiong2025vision, cheng2025open}, which yields robot-native data with minimal embodiment gap but is slow, costly, and difficult to deploy for mobile manipulators in diverse environments. 
UMI~\cite{chi2024universal, zhaxizhuoma2025fastumi} addresses scalability by enabling in-the-wild data collection with a portable handheld system.
While UMI minimizes the embodiment gap by using wrist-mounted cameras and relative end-effector control, its reliance on wrist-centric sensing fundamentally limits the observability of the global task context. 
%
Recent UMI extensions incorporate an external camera~\cite{rayyan2025mv} or VR headsets~\cite{zeng2025activeumi, yu2025egomi}, but their stationary setups or motion sickness limit their application to fixed-base tasks.
In contrast, \ours integrates a non-intrusive head-mounted camera into the UMI framework, enabling seamless and scalable deployment in dynamic mobile environments. 


\subsection{Robot Learning from Egocentric Demonstrations}
\vspace{-1mm}
Egocentric human demonstrations offer a scalable data source for learning bimanual manipulation.
Prior works leverage large-scale human videos~\cite{lepert2025masquerade, yang2025egovla, cai2025n} or utilize wearable devices for scalable data collection~\cite{kareer2025egomimic, punamiya2025egobridge, zhu2026emma, hoque2025egodex, qiu2025humanoid, punamiya2026egoverse}.
However, they still require co-training or fine-tuning with robot teleoperation data due to the large human-to-robot embodiment gap.
%
In addition to learning bimanual manipulation, recent works further leverage egocentric demonstrations to learn active perception behaviors~\cite{xiong2025vision, zeng2025activeumi, yu2025egomi, chuang2025look}. However, these approaches assume a robot with a customized 6-DoF neck to directly mimic human head motions, bypassing the kinematic and action-space gaps between human and robot heads.
On the contrary, we leverage a 3D visual representation and a look-at point action abstraction to transfer active perception behaviors from human demonstrations to a standard bimanual mobile manipulator with only a 2-DoF neck.

\subsection{Learning Mobile Manipulation From Demonstrations}
\vspace{-1mm}
Mobile manipulation couples long-range navigation with precise manipulation, making it challenging to learn from human demonstrations.
While learning decoupled navigation-manipulation strategies~\cite{xiong2024adaptive, uppal2024spin,yang2025mobi} simplifies the problem, these methods limit the ability to imitate end-to-end behaviors directly from human demonstrations. Recent works learn policies that predict end-effector commands, employing a whole-body controller to realize them through coordinated motion~\cite{ha2025umi, sundaresan2025homer}. While effective, these pipelines have primarily been demonstrated on \emph{single-arm} platforms.

Scaling to the \emph{bimanual} setting introduces distinct challenges, where two-arm coordination, base positioning, and active perception must be synchronized.
Although low-cost whole-body interfaces~\cite{fu2024mobile, dass2024telemoma, jiang2025behavior} attempt to ease the collection of such coordinated bimanual demonstrations, their dependence on robot teleoperation creates a bottleneck for data scalability. Alternative approaches explore in-the-wild data collection with wearable devices~\cite{zhu2026emma}, learning from human videos~\cite{bahety2025safemimic}, or automated data generation through simulation~\cite{li2025momagen}, yet these methods still require robot teleoperation data for fine-tuning. 
In contrast, \ours allows mobile manipulation directly from robot-free human demonstrations.

\section{Design Objectives}
\vspace{-1mm}
The goal of this paper is to design a general learning from demonstration framework for whole-body mobile manipulation for diverse manipulation tasks. To meet this requirement, we target the following system capabilities: 
\begin{itemize}[leftmargin=3mm]
    \item \textit{Scalability}: fast, intuitive, and portable demonstration interface for data collection in diverse environments.  
    \item \textit{Transferability}: overcoming both visual and kinematic embodiment gaps from human demonstrators to robots.
    \item \textit{Whole-body coordination}: able to efficiently coordinate whole-body action to realize both \textit{precise} end-effector tracking for accurate manipulation and \textit{effective} active perception to intentionally gather task-relevant information.  
\end{itemize}


Fig.~\ref{fig:pipeline} shows an overview of our system. We achieve scalability through an intuitive data collection interface (\S~\ref{sec:data_collection}), transferability through a cross-embodiment hand-eye policy trained on the collected  demonstrations (\S~\ref{sec:policy}), and whole-body motion through a  whole-body controller (\S~\ref{sec:wbc}) that executes policy outputs under the robot's physical constraints. 

\section{HoMMI Data Collection Interface}
\vspace{-1mm}
\label{sec:data_collection}
To enable scalable, robot-free demonstration data collection for bimanual mobile manipulation, we adapt the UMI gripper design while extending it with an egocentric view and head motion capture.
Concretely, the data collection system uses three iPhones: two mounted on the grippers and one mounted on a cap (Fig.~\ref{fig:pipeline} left). We leverage Apple’s ARKit multi‑device collaboration to establish a shared coordinate frame across phones.
During each demonstration, we record RGB video, depth maps, 6-DoF poses, and gripper widths at \SI{60}{Hz} on all three iPhones, producing synchronized multimodal trajectories that are directly consumable by our downstream policy learning pipeline (\S~\ref{sec:policy}).
The interface is designed to be intuitive and lightweight, providing direct visual and haptic feedback to the operator and avoiding the motion-sickness often associated with VR-based data collection~\cite{yu2025egomi, zeng2025activeumi, xiong2025vision}.


\section{Cross-embodiment Hand-Eye Policy}
\vspace{-1mm}
\label{sec:policy}
Leveraging the collected data, we train an end-to-end visuomotor policy based on Diffusion Policy~\cite{barreiros2025careful, chi2025diffusion}.
At each time step $t$, the policy conditions on a short observation window $O_t={o_{t-T_o+1},\ldots,o_t}$ and predicts a horizon of actions $A_t={a_{t+1},\ldots,a_{t+T_p}}$. However, naively adding the head RGB image to the observation and directly predicting the head pose as part of the action substantially enlarges the human-robot embodiment gap, often leading to failures in deployment. We therefore introduce three key algorithmic designs that overcome these transferability challenges, including (1) a 3D visual representation, (2) a 3D look-at point action representation, and (3) a gripper-centric frame shared by observations and actions. The center of Fig.~\ref{fig:pipeline} shows an overview of our policy.

\begin{figure}[t]
    \centering
    \includegraphics[width=0.9\linewidth]{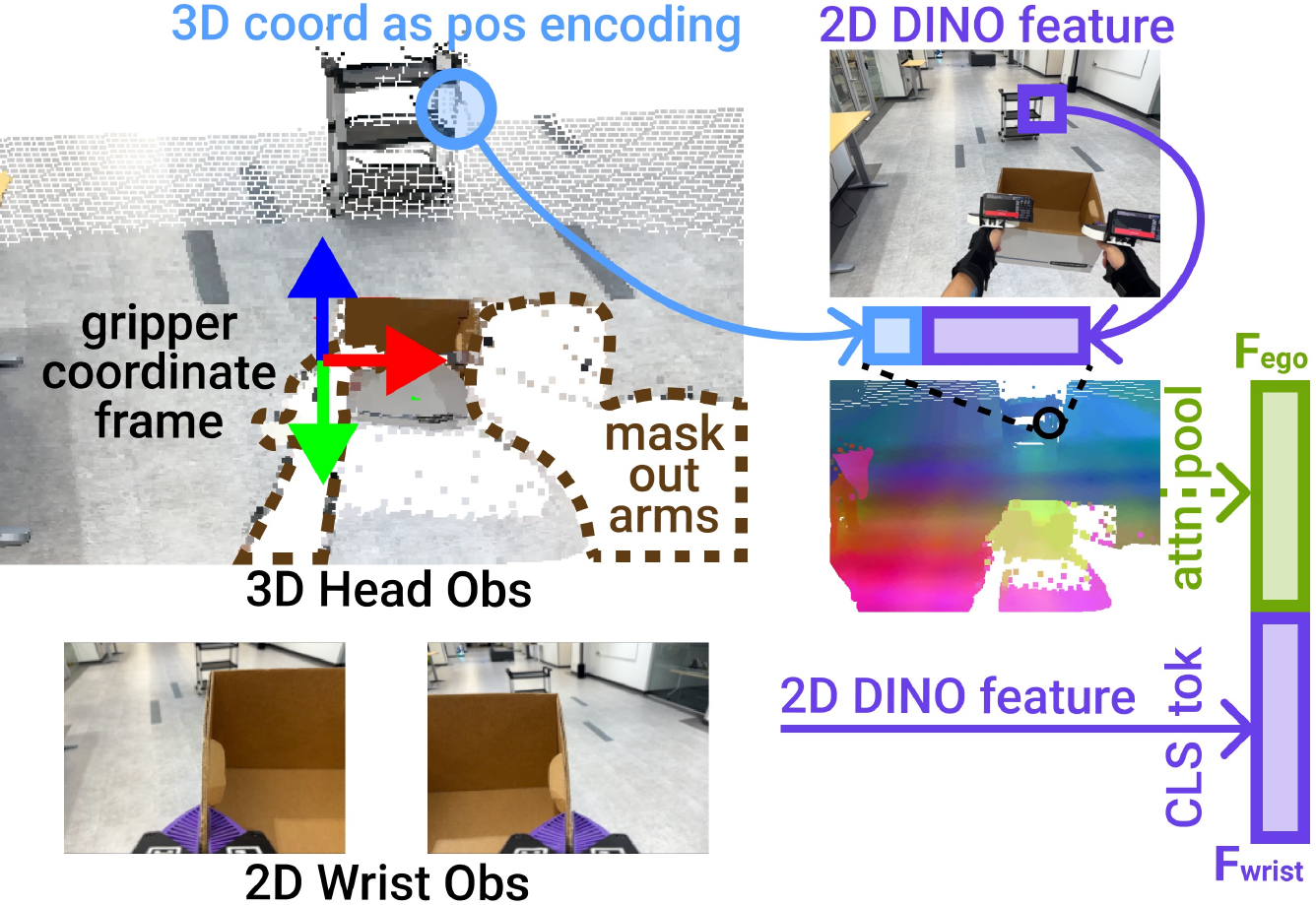}
    \caption{\textbf{Embodiment-Agnostic Visual Representation.} We use a 3D representation for egocentric observations that allows using an embodiment-agnostic gripper coordinate frame, and masking out embodiment-specific arms and body observations.}
    \label{fig:obs_repr}
    \vspace{-6mm}
\end{figure}
\subsection{3D Visual Representation to Mitigate the Visual Gap}
\vspace{-1mm}
Head-mounted RGB cameras often exhibit larger viewpoint and appearance differences between the human and robot compared to wrist-mounted cameras. Consequently, instead of directly feeding head RGB to the policy, 
we lift the egocentric observations into 3D and encode them with geometry‑aware tokens, inspired by Adapt3R~\cite{wilcox2025adapt3r}. Specifically, for each head camera frame, 
we first obtain a pointmap (from iPhone depth or stereo depth estimation~\cite{wen2025foundationstereo} on the robot), and patchify and downsample it via nearest neighbor interpolation s.t. each $16\times16$ patch corresponds to one 3D point. We then process the RGB frame by extracting a DINO-v3 ViT patch feature~\cite{simeoni2025dinov3, xu2023jacobinerf} for each patch. These patch features are further lifted to 3D by concatenating them with a sinusoidal encoding of the corresponding 3D point in the downsampled pointmap, tying appearance feature to 3D geometry and making the feature robust to head pose and height changes. To further reduce the appearance mismatch, we mask out arm points by transforming the pointmaps into left/right gripper frames and discarding points with $z<0$, since arms originate behind the grippers. In the end, we use an attention pooling layer to process all tokens and obtain a head observation embedding.


Fig.~\ref{fig:obs_repr} illustrates the visual representation of our policy. The entire observation embedding includes the 3D representation mentioned above, a 2D representation for wrist images, and proprioception. Concretely, we finetune a shared \texttt{dinov3-vitb16} encoder for wrist and head images. Wrist images are resized to $224\times224$ and represented by the CLS token features $F_{wrist}$. The egocentric image is resized to $512\times512$, split into $32\times32=1024$ image patches, augmented with 3D positional encoding, and downsampled to 512 tokens; attention pooling (with the arm attention mask) yields $F_{ego}$.

\subsection{3D Look-at Point Action Representation to Mitigate the Kinematic Gap}
\label{sec:lookat_point}
\begin{wrapfigure}{r}{.51\linewidth}
    \centering
    \vspace{-4mm}
    \includegraphics[width=\linewidth]{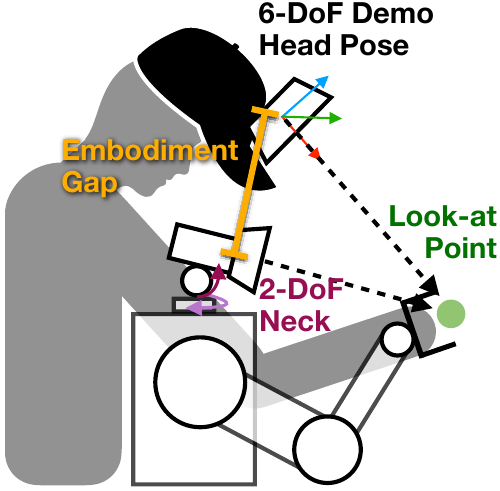}
    \caption{
    \textbf{Look-at Point Action Representation.} To bridge the kinematic gap (e.g., height and neck DoF), we relax the head action constraint by representing the robot gaze as a ``3D look-at point''. This representation allows effective active perception for gathering task-relevant information without over-constraining the robot to mimic human head motions exactly. 
    }
    \label{fig:look-at}
\end{wrapfigure}

\begin{figure*}[t]
    \centering
    \includegraphics[width=\linewidth]{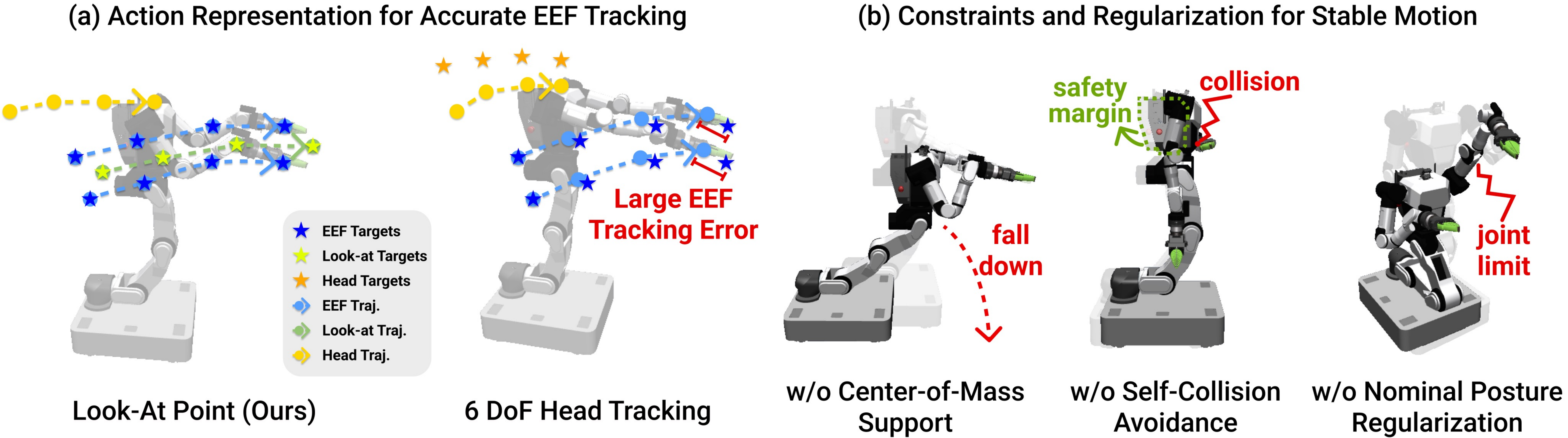}
    \caption{\textbf{\ours Whole-Body Controller} is designed to achieve \textit{precise} end-effector tracking for accurate manipulation and \textit{effective} active perception for information gathering. To do so, it uses (a) a relaxed head look-at point action representation that allows accurate bimanual end-effectors SE(3) tracking, circumventing the infeasibility and increased error associated with simultaneous 6-DoF head-hand tracking.
    In addition, we also apply (b) constraints and regularization to ensure stability and prevent the disastrous behaviors that would otherwise occur.
    }
    \label{fig:wbc}
    \vspace{-6mm}
\end{figure*}

Mobile robots have different kinematics than human demonstrators (e.g., shorter torso and fewer degrees of freedom in the neck). As a result, directly mimicking 6‑DoF head poses from human data can easily produce infeasible motions. 
We instead control head motion via a \emph{3D look‑at point} $\ell_t\in\mathbb{R}^3$ (Fig.~\ref{fig:look-at}). This relaxed representation preserves active perception intent while respecting kinematic constraints (Fig.~\ref{fig:wbc}a).


During training, the look‑at point is computed as the intersection of the center camera ray with the scene pointmap. 
At inference, the head controller converts $\ell_t$ to a feasible head orientation by constructing a rotation whose forward axis points toward $\ell_t$. Let $c_t\in\mathbb{R}^3$ be the current head position and let $R_t^{\text{cur}}=\begin{bmatrix}
x_t & y_t & z_t
\end{bmatrix}\in \mathbb{R}^{3\times 3}$ be the current head orientation, where $x_t$ denotes the current head $x$‑axis. We define the desired viewing direction as a unit vector pointing from the current position to the look-at point,
\(
\hat{d}_t=\frac{\ell_t-c_t}{\|\ell_t-c_t\|}. 
\)
We then project the current $x$‑axis onto the plane orthogonal to $\hat{d}_t$,
\(
x_t' = x_t - (x_t^\top \hat{d}_t)\hat{d}_t,
\hat{x}_t=\frac{x_t'}{\|x_t'\|},
\)
and construct the remaining axis
\(
\hat{y}_t=\hat{d}_t \times \hat{x}_t.
\)
The target head rotation is then
\(
R_t=\begin{bmatrix}\hat{x}_t & \hat{y}_t & \hat{d}_t\end{bmatrix}.
\)
If $\|x_t'\|$ is near zero, we replace $x_t$ with a fixed world‑up vector before projection.
This yields a feasible head command without constraining the policy to robot‑specific pose limits.


\subsection{Gripper-Centric Frame for Spatial Awareness}
\vspace{-1mm}
In our system, hand-eye coordination requires a reference frame that keeps observations and actions in-distribution. Egocentric frames shift with head motion and embodiment differences (height, neck DoF, camera placement), which hurts transfer from human demonstration to robot.
We therefore express all observations and actions in a gripper‑centric frame by transforming gripper poses (both proprioception and action), head pointmaps, and look‑at points to the left-gripper frame, so the policy always reasons in a consistent spatial frame centered at the manipulator. This anchors observation and action to the manipulators that execute the task, improving spatial awareness for 3D representations and reducing cross-embodiment mismatch compared to an egocentric frame that drifts with out-of-distribution~(OOD) head motion.

\section{Robot System}
\vspace{-1mm}
\subsection{Bimanual Mobile Manipulator Hardware Setup}
\label{sec:robot_hardware}

\begin{wrapfigure}{r}{.5\linewidth}
    \centering
    \includegraphics[width=\linewidth]{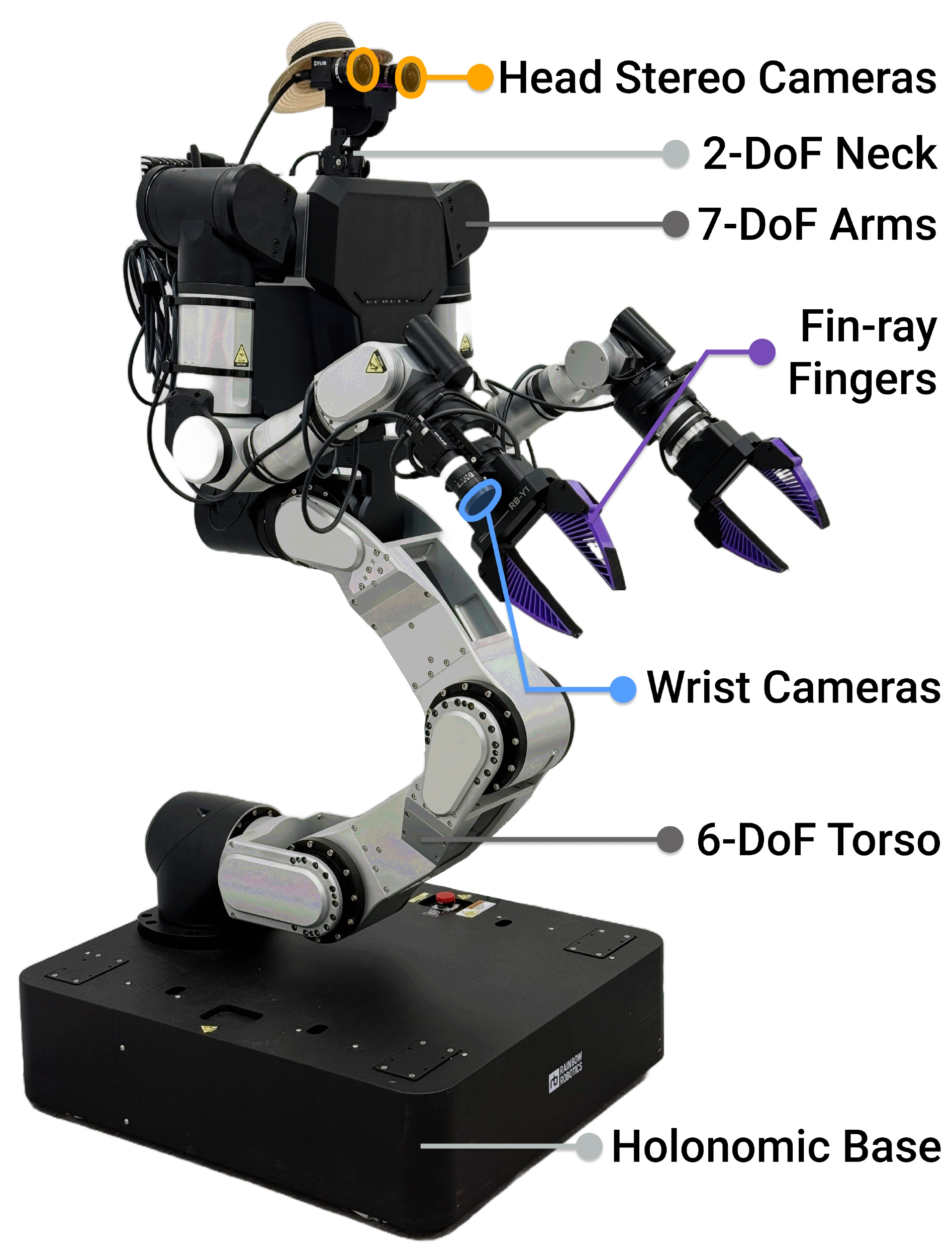}
    \caption{
    \textbf{\ours Robot Hardware} features a high DoF bimanual mobile manipulator with customized cameras and fingers that match the \ours data collection hardware.
    }
    \label{fig:robot}
    \vspace{-2mm}
\end{wrapfigure}

We build a mobile bi-manipulation platform targeting generalizability, observability, and transfer- ability of the learned policy (Fig.~\ref{fig:robot}). 
We employ the \texttt{Rainbow Robotics RB-Y1} as a core platform, equipped with two 7-DoF arms and a 6-DoF torso on a holonomic base to support diverse mobile manipulation tasks. It also supports active perception via a 2-DoF neck, on which we install a stereo pair of industrial-grade wide-angle cameras (\texttt{FLIR BFS-PGE-23S3C-CS}) to capture egocentric context. 
To align training and deployment setup, we mount fin-ray fingers identical to the UMI grippers on the end-effectors and mount wrist-mounted cameras (\texttt{FLIR BFS-PGE-50S5C-C}) at similar locations.

\subsection{Constraint-Aware Whole-body Controller}
\vspace{-1mm}
\label{sec:wbc}
Since our policy outputs end-effector poses and head look-at points, we need a whole-body controller to solve whole-body joint actions and base motions to achieve the cartesian space end-effector trajectory commands. Specifically, the whole-body controller needs to meet these requirements: accuracy (low tracking error), smoothness (non-jerky motion), stability (no falls or self-collisions), and human-likeness (similar range of motion as the demonstrator).

    
    

To satisfy these requirements, we implement a differential whole-body IK solver using \texttt{Mink}~\cite{Zakka_Mink_Python_inverse_2025} with (i) high-weight bimanual $\mathrm{SE}(3)$ tracking terms to prioritize accuracy, (ii) temporal command interpolation combined with posture and velocity regularization to encourage smooth motions, (iii) explicit constraints and tasks such as torso upright orientation, center-of-mass (CoM) support, and self-collision avoidance, to ensure stability; and (iv) regularization toward a nominal ``human'' posture and a balanced allocation between arm motion and base motion to produce human-like behavior (Fig.~\ref{fig:wbc}b).

Concretely, let $\Delta q\in \mathbb{R}^{n_v}$ be the velocity DoFs, define the objective function 
$f(\Delta q)=C_\text{ee}(\Delta q)+C_\text{nominal}(\Delta q)+C_\text{current}(\Delta q)+C_\text{com}(\Delta q)$.
The costs include (1) $C_\text{ee}$ end-effector pose tracking (primary task); (2) $C_\text{nominal}$ a nominal posture task to bias toward a preset human-like configuration; (3) $C_\text{current}$ a current posture task to discourage sudden posture changes; and (4) $C_\text{com}$ a CoM-over-base task to keep the body mass supported by the base. At each timestep, we solve for $\Delta q$ using a constrained quadratic program, 
\[\begin{aligned}
\min_{\Delta q \in \mathbb{R}^{n_v}} \quad & f(\Delta q) + \lambda \left\|\Delta q\right\|_2^2\\
\text{s.t.}\quad 
& G_{\text{cfg}} \Delta q \le h_{\text{cfg}} \\
& G_{\text{joint-vel}} \Delta q \le h_{\text{joint-vel}} \\
& G_{\text{base-vel}} \Delta q \le h_{\text{base-vel}} \\
& G_{\text{coll}} \Delta q \le h_{\text{coll}} \\
& A_{\text{upright}} \Delta q = 0
\end{aligned}\]
where $\lambda$ is the damping coefficient. The inequality constraints $G_j\Delta q\le h_j$ encode configuration bounds $G_{\text{cfg}}$, joint velocity bounds $G_{\text{joint-vel}}$, base velocity bounds $G_{\text{base-vel}}$, and collision avoidance limits $G_{\text{coll}}$. 
Finally, the equality constraint $A_{\text{upright}} \Delta q = 0$ enforces a zero-sum constraint on the three torso joints for an upright posture. 
Together, these tasks and constraints balance accuracy, smoothness, stability, and human-likeness in a single optimization.
Importantly, the look-at point is controlled separately from the whole-body IK and is excluded from the IK objective to avoid interfering with bimanual EEF tracking. As described in Sec.~\ref{sec:lookat_point}, the neck orientation is computed to point toward the policy-predicted look-at point from the robot’s current head position.

We run this IK solver asynchronously at \SI{100}{Hz} to bridge the \SI{10}{Hz} policy loop and the \SI{500}{Hz} robot control loop. The policy produces a stream of target end-effector poses with specified command durations (\SI{0.1}{s}). At each IK tick, we compute an interpolated target by linearly blending the previous and current targets based on the elapsed fraction of the command duration. This reduces discontinuities at policy update boundaries and improves tracking smoothness.

\subsection{Asynchronous Policy Inference}


Mobile manipulation cannot be paused for inference without introducing base jerks and tracking errors, so we decouple perception, policy inference, and whole-body control. 
We run a \emph{detached policy server} that receives timestamped observations, performs inference, and returns a timestamped action chunk; and a \emph{real-time execution bridge} that aligns observations across sensors to prepare timestamped observations for the policy, receives actions from the policy, filters stale actions, and streams time‑aligned targets to the whole‑body controller.

At each inference cycle, the bridge collects a history of camera frames and proprioception, corrects each camera stream by a measured latency, and then uses the \emph{latest} camera timestamp as the anchor. Proprioception is interpolated to these anchor timestamps. This yields a \textit{synchronized observation window}, similar to the latency matching in UMI~\cite{chi2024universal}. 
The policy server takes in these observations, runs policy inference, and outputs a horizon of actions whose timestamps are anchored to the last observation time.
The bridge then discards any actions whose timestamps fall \emph{before} the earliest feasible execution time, given inference time and execution latency, and updates the \textit{scheduled action buffer} with the remaining actions.
%
These buffered actions are streamed at \SI{10}{Hz} to provide latency-matched targets to the whole-body controller.
\begin{figure*}[t]
    \centering
    \includegraphics[width=\linewidth]{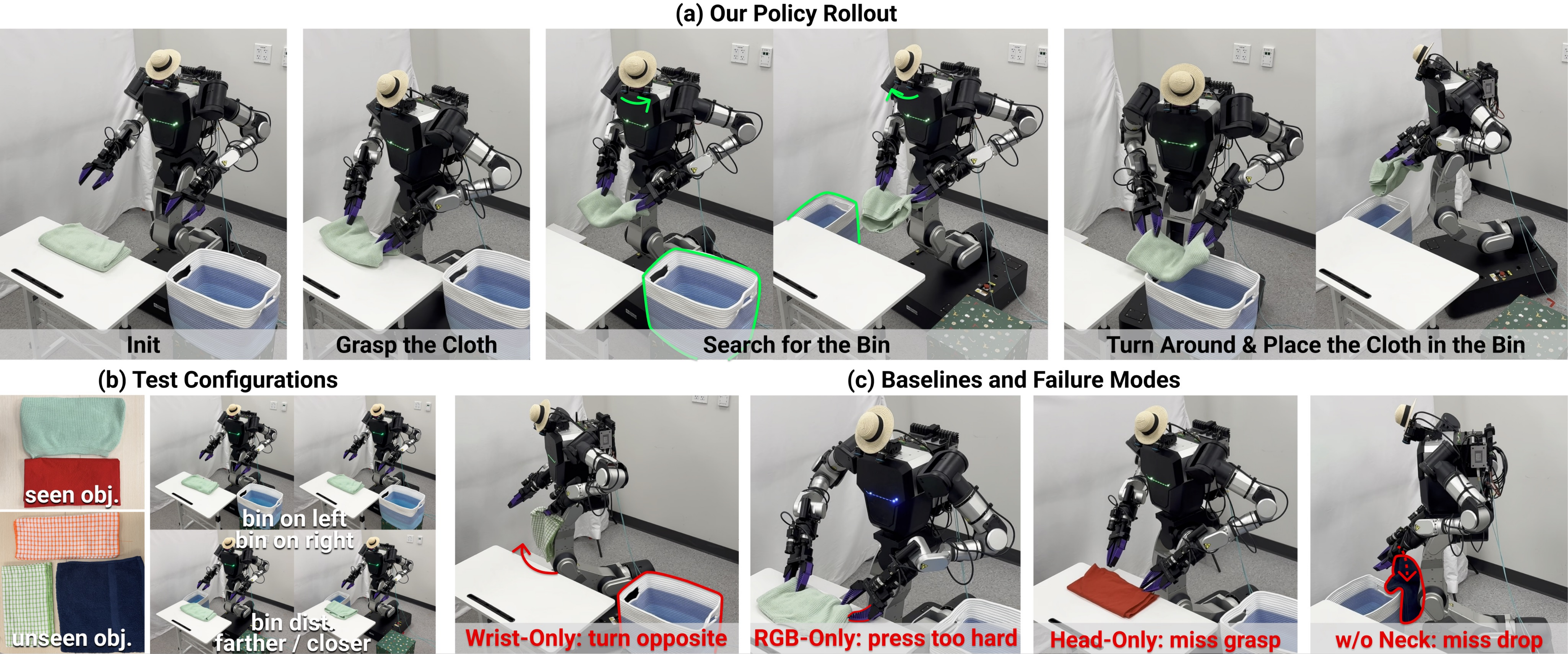}
    \caption{\textbf{Laundry Task.} (a) Our cross-embodiment hand-eye policy rollout, highlighting our system's capability of whole-body coordination and active perception. (b) Different test scenarios with different objects and bin locations. (c) Typical failure cases of the baselines. }
    \label{fig:laundry_result}
    \vspace{-5mm}
\end{figure*}

\begin{figure}[t]
    \centering
    \includegraphics[width=\linewidth]{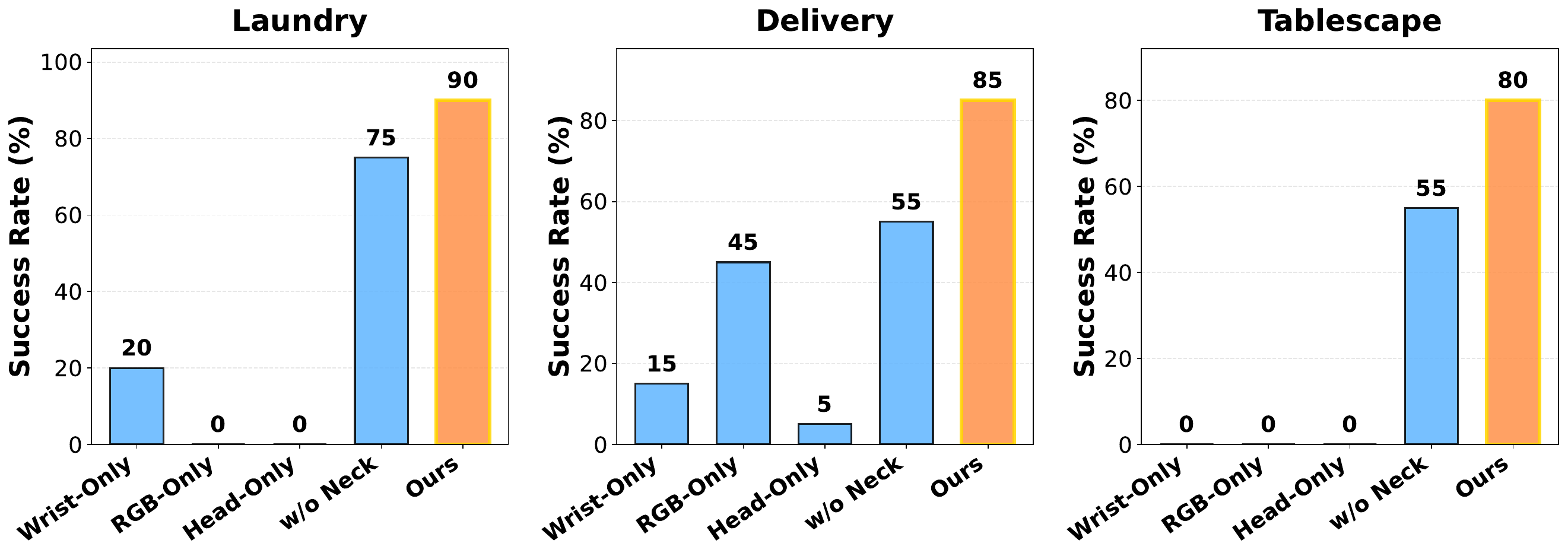}
    \caption{\textbf{Quantitative Results.} \texttt{Ours} consistently outperforms baselines across all three long-horizon mobile manipulation tasks.}
    \label{fig:plot}
    \vspace{-7.5mm}
\end{figure}

\section{Evaluation}

We evaluate whether long-horizon bimanual mobile manipulation can be learned \emph{directly} from robot-free human demonstrations and transferred to a real mobile manipulator. 
Specifically, our evaluation probes four core capabilities:
\begin{itemize}[leftmargin=3mm]
    \item \textbf{Cross-embodiment transfer}: deploying policies learned from robot-free human demonstrations on a robot with a different appearance and kinematics. Required for all tasks.
    \item \textbf{Bimanual / Whole-body coordination}: coordinating two arms, mobile base, torso, and head for mobile manipulation.
    \item \textbf{Long-horizon navigation}: moving through a large workspace and approaching targets whose locations vary.
    \item \textbf{Active perception}: intentionally controlling head motion to acquire task-relevant information that may initially be outside the field of view.
\end{itemize}
We compare \ours to the following baselines and ablations:
\begin{itemize}[leftmargin=3mm]
    \item \texttt{Wrist-Only(UMI)}: the original UMI~\cite{chi2024universal, ha2025umi} setup, using only wrist RGBs as input and gripper trajectories as output.
    
    \item \texttt{RGB-Only(UMI+Ego)}: naively adding head RGB to the UMI design and predicting gripper and 6-DoF head actions directly. This setup is similar to ViA~\cite{xiong2025vision}, however, we use a wearable UMI device for data collection instead of teleoperating the same robot embodiment, which provides better scalability but also introduces additional challenges in cross-embodiment policy learning. 
    
    \item \texttt{Head-Only}: removing wrist RGBs from \texttt{Ours} policy observation and only using the 3D head observation.
    \item \texttt{w/o Active Neck}: running \texttt{Ours} policy but disabling head motion control.
\end{itemize}

\begin{figure*}[t]
    \centering
    \includegraphics[width=\linewidth]{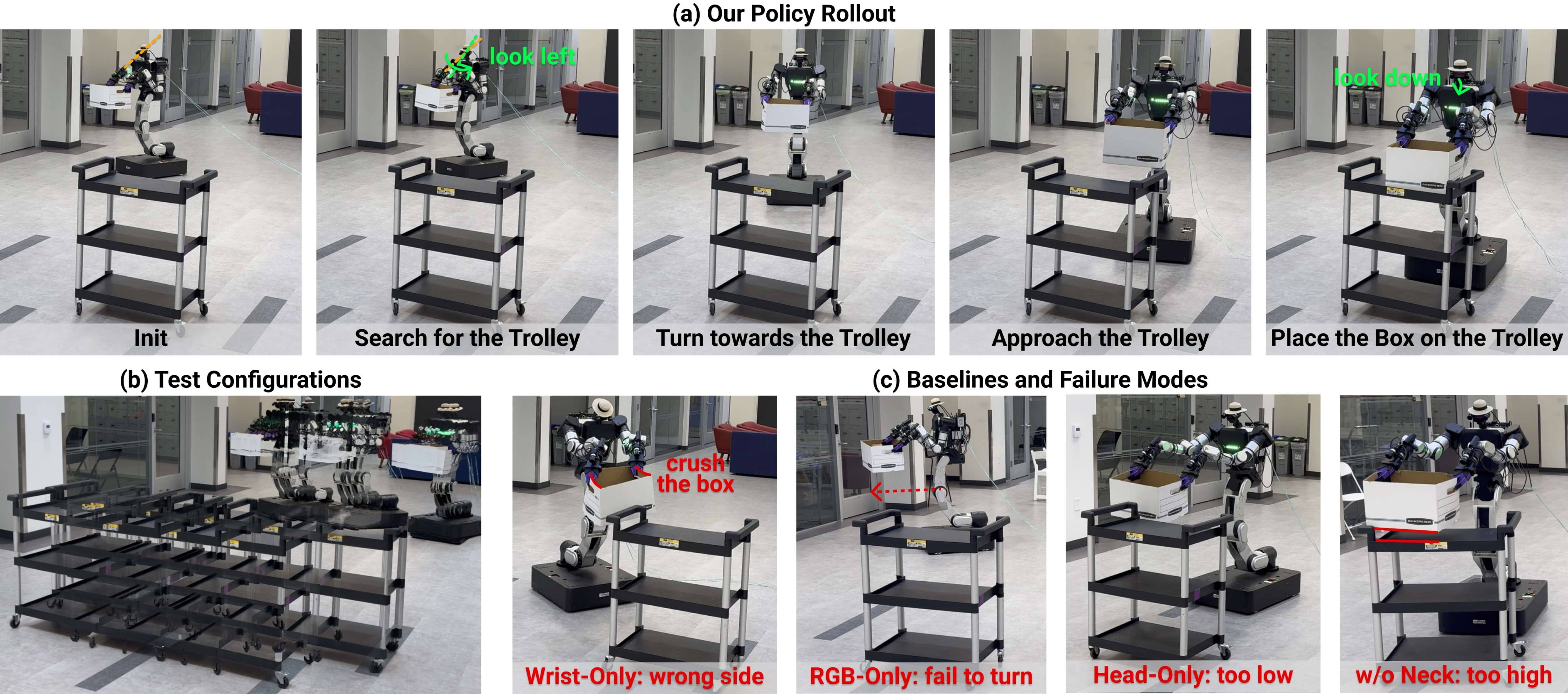}
    \caption{\textbf{Delivery Task.} (a) Our policy rollout, demonstrating long-horizon navigation over a large workspace and active perception. (b) Different test scenarios with different trolley locations and initial base positions and orientations. (c) Typical failure cases of the baselines.}
    \label{fig:delivery_result}
    \vspace{-5mm}
\end{figure*}

\vspace{-1mm}
\subsection{Laundry Task}
\vspace{-1mm}
\noindent \textbf{Task:} As shown in Fig.~\ref{fig:laundry_result}a, the robot approaches a table, grasps a cloth with both hands, searches for a bin, navigates, and places the cloth in the bin. The task success rate is measured by whether the cloth is placed in the bin in the end.

\noindent \textbf{Capability:} 
\underline{Bimanual coordination:} The robot must grasp the cloth firmly with both hands.
\underline{Whole-body coordination:} The bin is placed to the side and lower than the table, thus requiring the robot to flexibly coordinate whole-body motion to navigate, rotate, and bend down to approach the bin and place the cloth into it.
\underline{Active perception:} The bin may be outside the camera view after grasping, thus requiring the robot to actively search for it by looking sideways until it locates it.

\noindent \textbf{Data Collection:} We collected 200 demonstrations with randomized bin locations, initial configurations, and objects.

\noindent \textbf{Test Scenarios:}
As shown in Fig.~\ref{fig:laundry_result}b, we ran evaluation across a total of 20 rollouts, involving 5 objects (2 seen and 3 unseen) and 4 bin configurations (2 on the left and 2 on the right).

\noindent \textbf{Performance:}
Quantitative results are shown in Fig.~\ref{fig:plot} (left).
\texttt{Ours} achieves a 90\% success rate. It flexibly coordinates whole-body motion to navigate the workspace and place the cloth in the bin, robustly searches the environment to find the bin, and always turns correctly. Occasional failures result from not grasping the cloth firmly enough, causing it to slip halfway. 

Fig.~\ref{fig:laundry_result}c shows the baselines' typical failure modes. (1)~\texttt{Wrist-Only} policy's dominant failure is consistently turning to one side, regardless of the bin location, due to the bin not being visible from the wrist camera view. 
Other failure cases include not grasping firmly enough and inaccurately placing the cloth in the bin. 
We hypothesize that these issues are due to the lack of global context and spatial information from wrist views.
(2)~\texttt{RGB-Only} consistently fails to grasp and presses the table too hard, triggering the robot's wrench safety guard, which we hypothesize is due to egocentric RGB having appearance and viewpoint mismatches in human and robot observations, causing the policy to go OOD. 
%
(3)~\texttt{Head-Only}'s success rate is also 0\%, failing due to missing the cloth when attempting to grasp it, and only grasping one edge which leads to slip off later. Compared with \texttt{Ours}, this demonstrates that wrist cameras help provide local contact information that can improve grasping accuracy.
(4)~\texttt{w/o Active Neck} achieves a 75\% success rate, mostly failing to accurately place the cloth into the bin. We hypothesize that the lack of active perception causes the view to be more OOD and the bin to be not fully in view.

\vspace{-1mm}
\subsection{Delivery Task}
\vspace{-1mm}
\noindent \textbf{Task:} As shown in Fig.~\ref{fig:delivery_result}a, the robot carries a box, searches for a trolley, navigates over a large workspace, and places the box onto the trolley. 
The task success rate is measured by whether the box is eventually placed onto the trolley.  

\noindent \textbf{Capability:} 
\underline{Bimanual coordination:} Two hands need to maintain a stable distance to avoid crushing or tearing the box. They also need to coordinate heights to lift the box up and then lower it for accurate placement.
\underline{Long-horizon navigation:} The robot needs to navigate a large workspace (6$\times$6\SI{}{m}) and accurately approach the trolley in randomized locations.
\underline{Active perception:} The trolley may initially be out of view when the robot is rotated to face the other way, requiring the robot to search for the trolley, rotate, and then navigate over.

\noindent \textbf{Data Collection:} We collected 166 demonstrations with varying trolley locations and initial standing locations. 

\noindent \textbf{Test Scenarios:} As shown in Fig.~\ref{fig:delivery_result}b, We conducted 20 rollouts in total, consisting of 5 trolley locations and 4 different initial robot base initializations (position + yaw).

\noindent \textbf{Performance:}
Quantitative results are shown in Fig.~\ref{fig:plot} (middle).
\texttt{Ours} achieves 85\% success. The policy robustly performs visual servoing and long-horizon navigation, always approaching the correct direction towards the trolley. It also reactively adjusts the robot's approaching direction midway if the initial alignment is inaccurate. The remaining failures are due to slight misalignment at the end after long navigation.

Typical baseline failure modes are shown in Fig.~\ref{fig:delivery_result}c.
(1)~\texttt{Wrist-Only} achieves 15\%, the policy frequently approaches from an incorrect side or misaligns during placement, demonstrating that navigation and approach require global context beyond wrist views.
(2)~\texttt{RGB-Only} achieves 45\%. The policy consistently fails to turn towards the trolley when it is initially out of view because 6-DoF active head motion commands are unachievable by the whole-body IK due to kinematic infeasibility.
(3)~\texttt{Head-Only} achieves 5\%, often colliding the box with the trolley because the gripper heights are too low. This highlights that egocentric context alone is insufficient for manipulation precision.
%
(4)~\texttt{w/o Neck} achieves 55\%, often lifting the box too high during final placement due to the lack of a look down head motion.

\begin{figure}
    \centering
    \includegraphics[width=\linewidth]{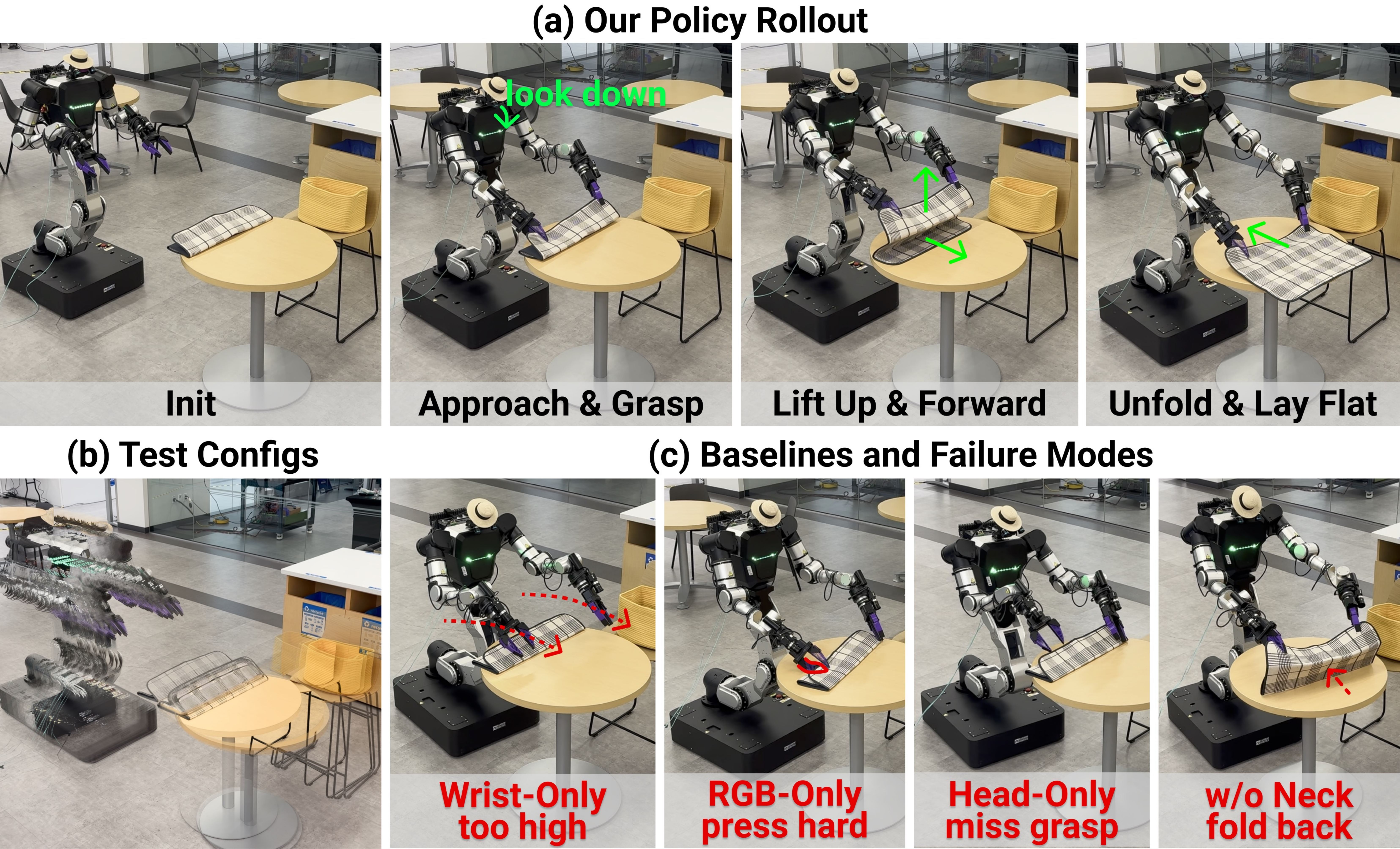}
    \caption{\textbf{Tablescape Task.} (a) Our policy rollout, demonstrating precise bimanual and whole-body coordination. (b) Different test scenarios with different initial base positions and mat placement. (c) Typical failure cases of the baselines.}
    \label{fig:table-result}
    \vspace{-5mm}
\end{figure}

\vspace{-1mm}
\subsection{Tablescape Task}
\vspace{-1mm}
\noindent\textbf{Task:} As shown in Fig.~\ref{fig:table-result}a, the robot approaches a table and grasps the two edges of a mat, carefully lifts the mat up and moves forward to unfold it, and finally lays the mat completely flat on the table and retracts its hands.

\noindent\textbf{Capability:} \underline{Bimanual coordination}: two hands need to precisely coordinate rotation and height to grasp the edges of the mat, and consistently maintain a stable distance and height to unfold it.
\underline{Whole-body coordination}: The robot needs to coordinate the base, torso, and arm motions to navigate across the workspace and significantly adjust the height of the grippers throughout the task.

\noindent\textbf{Data Collection:} We collected 115 demonstrations with varying initial standing locations and mat placements on the table.

\noindent\textbf{Test Scenarios:}
We ran 20 rollouts in total, including 5 initial base initializations and 2 mat configurations, and tried twice for each configuration (Fig.~\ref{fig:table-result}b).

\noindent\textbf{Performance:}
Quantitative results are shown in Fig.~\ref{fig:plot} (right).
\texttt{Ours} achieves an 80\% success rate, demonstrating robust recovery behaviors. When the mat is not perfectly aligned with the table or folds back on the first attempt, the robot lifts the mat again and retries until it successfully unfolds. Occasional failure cases arise from slightly missing the grasp. 

Fig.~\ref{fig:table-result}c shows the failure modes of the baselines.
(1)~\texttt{Wrist-Only} grippers rotate too late and go well above the mat, which we hypothesize is due to the lack of global spatial context.
(2)~\texttt{RGB-Only} presses the grippers too hard against the table, potentially due to OOD head observations.
(3)~\texttt{Head-Only} misses contact with the mat, demonstrating the need for wrist cameras to provide local contact information.
(4)~\texttt{w/o Neck} achieves 55\% success, with failures resulting from missing the grasp and failing to recover when the mat folds back, potentially due to the inability to actively adjust the viewpoint for better alignment observation.

\subsection{Findings Summary}
\begin{figure}[t]
    \centering
    \includegraphics[width=\linewidth]{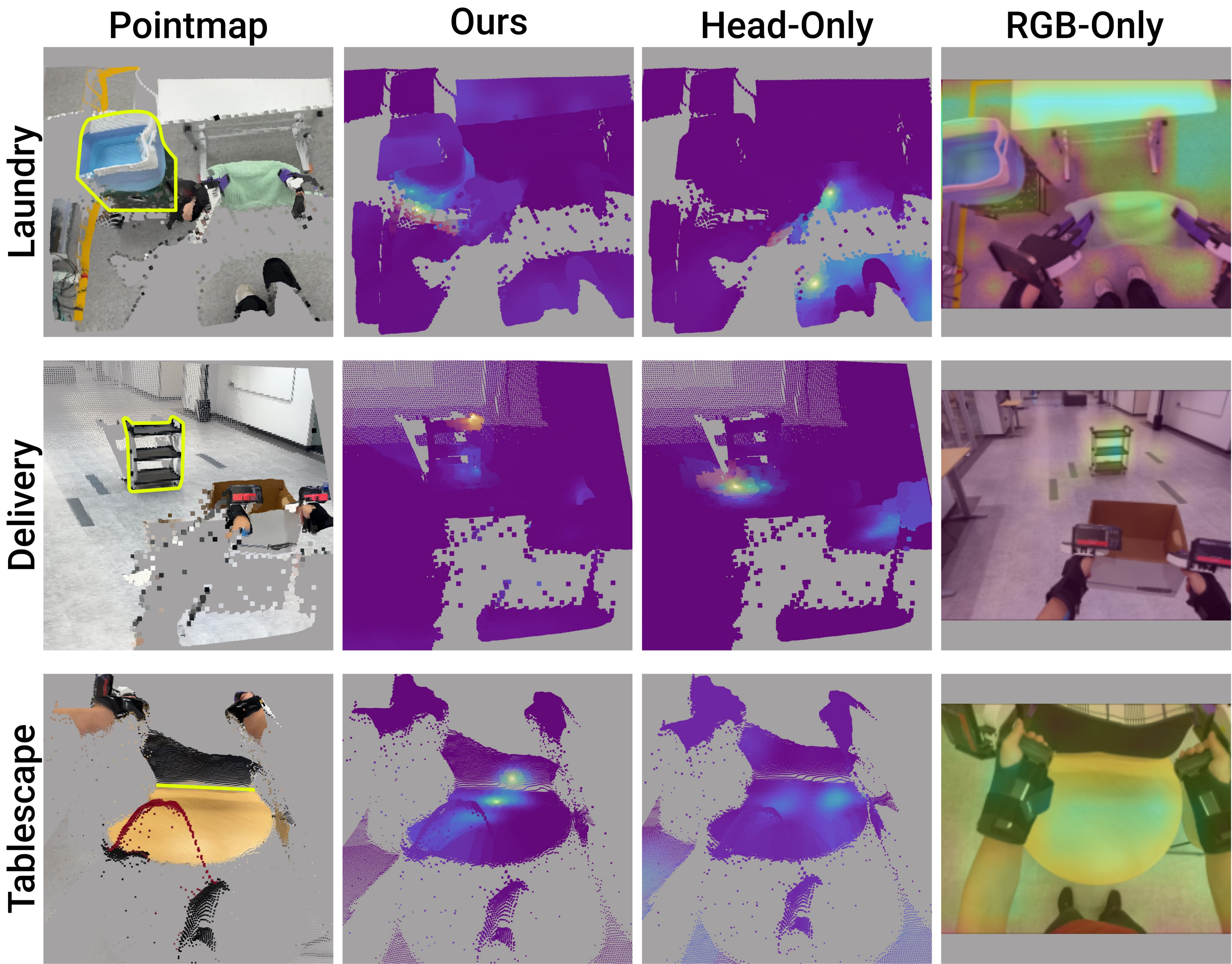}
    \caption{\textbf{Egocentric Attention Comparison.} We visualize attention maps for egocentric observations with yellow representing higher attention values. \texttt{Ours} exhibits clean attention highlighted around task-relevant objects, while baselines' attentions are less informative. 
    }
    \label{fig:attn}
    \vspace{-6mm}
\end{figure}

\noindent\textbf{F1: Wrist-only sensing under-observes global task context and bimanual coordination.} 
\texttt{Wrist-Only} policy exhibits poor performance on all tasks that require search, navigation, and alignment, which depend on the wider scene. 
It is not capable of actively searching for task-relevant context in the scene due to its limited field of view. 
It is brittle in long-horizon navigation, drifting easily and unable to recover from failures due to the lack of global task progress.
It also lacks spatial awareness of the other hand, which causes failures in coordinating both hands for precise bimanual manipulation.
On the contrary, \ours augments UMI with \emph{egocentric sensing}, providing global context and active perception behaviors crucial for mobile manipulation.

\noindent\textbf{F2: Head-mounted camera alone is insufficient.}
While being the most common camera configuration for humanoid design~\cite{cheng2025open, pan2025lookout, fu2025humanplus}, the \texttt{Head-Only} baseline that relies solely on a head-mounted camera fails in grasping and alignment.
\ours combines head camera views with wrist camera views, which provide essential local contact cues for fine-grained manipulation.
We also find that having wrist visual observations as policy input and jointly finetuning the vision encoder on both wrist and egocentric images helps the policy learn cleaner egocentric attention that is more focused on task-relevant objects (Fig.~\ref{fig:attn} \texttt{Ours} v.s. \texttt{Head-Only}).

\noindent\textbf{F3: Naively adding egocentric RGB can degrade performance under embodiment mismatch.}
Directly feeding the head RGB to the policy and regressing the head motion leads to brittle grasping and unstable motions, yielding a 0\% success rate on two tasks, indicating a significant OOD shift due to viewpoint/appearance mismatch. Tracking 6-DoF head and hand trajectories together often leads to large tracking errors and violates the robot's kinematic constraints (Fig.~\ref{fig:wbc}).
\ours bridges the visual gap by leveraging an embodiment-agnostic 3D egocentric visual representation, and bridges the kinematic gap through a relaxed 3D look-at point head action representation that allows the whole-body controller to achieve precise end-effector tracking and effective active perception.

\noindent\textbf{F4: Active head control effectively gathers task-relevant information and maintains policy observability.}
Disabling head motion reduces success, particularly when it is required to actively search for the object and precisely place or align objects.
This supports that our look-at point based active head control \emph{effectively} imitates human's active perception behavior to gather task-relevant information and aligns the egocentric view more closely with the training distribution.

\noindent\textbf{F5: Our cross-embodiment hand-eye policy learns task-relevant attention.}
As shown in Fig.~\ref{fig:attn}, \texttt{Ours} yields egocentric attention maps highlighted on task-relevant objects and contacts, demonstrating the effectiveness of our observation representation and gripper coordinate frame design. This also potentially helps to mitigate the visual embodiment gap, as the policy attends more to task-relevant regions than OOD observation points.

\noindent\textbf{F6: Our system allows transfer from human demonstrators with different body proportions.}
Demonstrations were collected from two operators with different heights (167\,cm and 182\,cm) and both transferred successfully. This supports that our 3D look-at point action representation mitigates the height difference among operators. 
\vspace{-1mm}
\section{Conclusion}
\vspace{-1mm}
We present \ours, a system that enables learning long-horizon whole-body mobile manipulation skills directly from robot-free human demonstrations. We employ a scalable data collection interface that augments bimanual UMI with egocentric sensing. To bridge the human-to-robot embodiment gap induced by egocentric sensing, we propose a cross-embodiment policy design with an embodiment-agnostic visual representation and a relaxed look-at point head action representation. A whole-body controller then achieves precise end-effector tracking and effective active perception by coordinating the robot's whole-body motions while respecting physical constraints.
Extensive real world experiments demonstrate that \ours allows versatile and challenging mobile tasks.
We hope the proposed data collection hardware, learning framework, and robot system setup will encourage and facilitate future research on democratizing and scaling mobile manipulation.

\section{Limitations and Future Work}
Our policy uses a short observation history, which can limit recovery in long-horizon tasks and under partial observability; incorporating longer-term memory~\cite{sridhar2025memer, henschel2025streamingt2v} is a natural next step. 
We rely on vision-only sensing; adding force/tactile sensing and explicit compliance control could improve contact-rich manipulation and safety~\cite{choi2026wild, liu2025vitamin, zhu2025touch, liu2023enhancing}. 
Finally, while we align camera placement and gripper geometry between data collection and deployment, there remains a hardware embodiment gap; better co-design, including generative hardware design~\cite{xu2025dynamics, kim2025generative}, may further improve policy transferability.

\section{Acknowledgments}
The authors would like to thank Calder Phillips-Grafflin, Aimee Goncalves, and Andrew Beaulieu from TRI for their help with the RB-Y1 hardware setup. 
Austin Patel for his help with the iPhone data collection app, Maximilian Du for his help with the fisheye camera calibration, and his feedback on the manuscript.
Kosei Tanada, Vitor Guizilini, Paarth Shah, Eric Dusel, Sam Creasey, Hillel Hochsztein, Benjamin Burchil, Mengchao Zhang, Mark Zolotas, and Naveen Kuppuswamy from TRI for their helpful discussions. 
Phoebe Horgan, Allison Henry, Richard Denitto, Maya Angeles, Owen Pfannenstiehl, Mariah Smith-Jones, and Gordon Richardson from TRI for their help with data collection.
Yifan Hou, Huy Ha, Hojung Choi, all REALab members, and Jinghan Sun for their helpful discussions and feedback on the manuscript.
This work was supported in part by the  NSF Award \#2143601, \#2037101, and \#2132519, and Toyota Research Institute. Dian Wang is supported by the HAI PostDoc fellowship.
The views and conclusions contained herein are those of the authors and should not be interpreted as necessarily representing the official policies, either expressed or implied, of the sponsors.

\bibliographystyle{plainnat}
\bibliography{references}

\clearpage
\setcounter{section}{-1}
\section{Appendix}
\section{Policy Training Details}
\textbf{Observations and actions.}
We use a short observation history of $T_o{=}2$ steps and predict an action horizon of $T_p{=}32$ steps at \SI{20}{Hz} (downsampled from \SI{60}{Hz} demonstrations). Observations include left/right wrist RGB ($224{\times}224$), head RGB and pointmap ($512{\times}512$), and proprioception (end‑effector poses and gripper widths). Actions are 23‑dimensional, including 9-dimensional left/right gripper poses, a 3-dimensional look-at point, and gripper widths ($23=2\times9+3+2$). Each
gripper pose is represented by a 3-dimensional vector for
position and a 6-dimensional vector for orientation, using the
first two columns of the rotation matrix~\cite{zhou2019continuity}.

\textbf{Model.}
We use Diffusion Policy~\cite{chi2025diffusion} with a Diffusion Transformers backbone (DiT)~\cite{barreiros2025careful}. The diffusion model is conditioned on global observation embeddings and predicts noise with a DDIM scheduler~\cite{song2020denoising}. We use $100$ training timesteps, $16$ inference steps, input perturbation $0.1$, and train with $8$ diffusion noise samples per observation. The DiT uses embedding dimension $768$, depth $10$, $12$ heads, MLP ratio $4$, and RMS norm~\cite{zhang2019root}. The observation encoder finetunes \texttt{dinov3‑vitb16} backbones~\cite{simeoni2025dinov3} for wrist and head RGBs with shared weights.

\textbf{Optimization.}
We use AdamW~\cite{loshchilov2017decoupled} with a cosine learning rate schedule starting at a learning rate of $7.5{\times}10^{-5}$ for the diffusion model and $7.5{\times}10^{-6}$ for finetuning the vision backbone, weight decay $1{\times}10^{-6}$, and betas $(0.95, 0.999)$. We train our policy and all baselines for 500 epochs.

\section{Whole-body Controller Details}
\subsection{Temporal Command Interpolation}
To ensure smooth trajectory tracking despite the lower policy frequency, the controller performs temporal interpolation of the target commands at each IK iteration. Given a new 6-DoF pose command received from the policy loop at time $t_0$ with a duration $T$, we compute a progress fraction $\alpha(t) = \min(1, \frac{t - t_0}{T})$. We derive the intermediate pose by interpolating its components: the position $\mathbf{p}$ uses linear interpolation,
\begin{equation}
    \mathbf{p}_{\text{interp}}(t) = (1 - \alpha)\mathbf{p}_{\text{prev}} + \alpha \mathbf{p}_{\text{cmd}},
\end{equation}
whereas the orientation is interpolated using spherical linear interpolation between the previous and target rotations. This interpolation approach ensures that the high-frequency whole-body controller receives continuous targets, effectively eliminating jitter in the resulting motion.

\subsection{Whole-body IK}
Our differential whole-body IK solver maps the interpolated Cartesian targets to generalized joint velocity $\Delta q$ by solving a differential IK problem. We use the \texttt{daqp} solver to handle the optimization, with the damping coefficient $\lambda$ set to $10^{-6}$ for numerical stability. The objective function $f(\Delta q)$ consists of the following individual costs:

\begin{itemize}[leftmargin=*]
    \item \textbf{Bimanual $\mathrm{SE}(3)$ Tracking} ($C_{\text{ee}}$):
    We penalize the Cartesian tracking error of both end-effectors in differential form:
    \[
    C_{\text{ee}} = \sum_{i\in\{\mathrm{L},\mathrm{R}\}} \left\| \mathbf{J}_i \Delta q - \mathbf{v}_i \right\|_{\mathbf{W}_{\text{ee}}}^2,
    \]
    where $\mathbf{J}_i$ is the Jacobian and $\mathbf{v}_i$ is the Cartesian command, with $\mathbf{W}_{\text{ee}}$ being a diagonal matrix of positional weights $w_p$ and rotational weights $w_o$.

    \item \textbf{Nominal Posture Regularization} ($C_{\text{nominal}}$):
    We regularize the solution toward a nominal posture $q_{\text{nom}}$ to encourage human-like, consistent body configurations. This nominal posture is defined by the robot's initial pose, which is set to match the human posture observed during data collection:
    \[
    C_{\text{nominal}} = \left\| (q+\Delta q)-q_{\text{nom}} \right\|_{\mathbf{W}_{\text{nom}}}^2,
    \]
    where $\mathbf{W}_{\text{nom}}$ is a diagonal matrix whose entries are set to $w_{\text{nom,torso}}$ for torso joints and $w_{\text{nom,arm}}$ for arm joints.
    
    \item \textbf{Current Posture Regularization} ($C_{\text{current}}$):
    To ensure motion smoothing in velocity space, we penalize any deviation from the current joint configuration and mobile base pose:    
    \[
    C_{\text{current}} = \left\| \Delta q \right\|_{\mathbf{W}_{\text{curr}}}^2,
    \]
    where $\mathbf{W}_{\text{curr}}$ is a diagonal matrix whose entries are set to the weight $w_{\text{curr}}$ for all joints, and $w_{\text{base,pos}}$ and $w_{\text{base,ori}}$ for the base translational and rotational components, respectively.
    
    \item \textbf{CoM-over-base Support} ($C_{\text{com}}$):
    We maintain the upper-body mass centered over the mobile base through the following objective and relative $XY$ displacement constraints:
    \begin{equation*}
    \begin{aligned}
        C_{\text{com}} &= \left\| p^{xy}_{\text{torso}}(q+\Delta q)-p^{xy}_{\text{base}}(q+\Delta q)-r^{xy}_{\star} \right\|_{\mathbf{W}_{\text{com}}}^2 \\
        \text{s.t.} \quad &| p^{x}_{\text{torso}} - p^{x}_{\text{base}} - r^{x}_{\star} | \le b_x, \quad | p^{y}_{\text{torso}} - p^{y}_{\text{base}} - r^{y}_{\star} | \le b_y,
    \end{aligned}
    \end{equation*}
    where $\mathbf{W}_{\text{com}}$ is a diagonal matrix with weights $w_{\text{com}}$ applied to the $x$ and $y$ dimensions, and $b_x, b_y$ are the permitted displacement bounds.
\end{itemize}

In addition to these cost terms, the solver must satisfy several physical and safety constraints, expressed as $G_j \Delta q \le h_j$ and $A \Delta q = b$, to maintain the robot's operational integrity:
\begin{itemize}[leftmargin=*]
    \item \textbf{Configuration Bounds} ($G_{\text{cfg}}$):
    These constraints ensure the next state $q + \Delta q$ remains within the physical joint position limits specified in the robot model.

    \item \textbf{Joint Velocity Limits} ($G_{\text{joint-vel}}$):
    Joint velocities follow the limits specified in the model, scaled by a safety factor of $0.9$. Specifically, \texttt{torso\_0}, \texttt{torso\_4}, and \texttt{torso\_5} are fixed to zero velocity to maintain a consistent torso orientation.

    \item \textbf{Base Velocity Limits} ($G_{\text{base-vel}}$):
    The mobile base velocity is also subject to the $0.9$ safety scale, with maximum limits set to \SI{1.0}{\meter/\second} for translation and \SI{1.0}{\radian/\second} for rotation.

    \item \textbf{Collision Avoidance} ($G_{\text{coll}}$):
    Collision avoidance is enforced over selected geometry groups, including (\textit{base+torso+head}, \textit{arms}) and inter-arm pairs. The solver maintains a minimum separation $d_{\text{safe}} = \SI{0.01}{\meter}$, activated within an influence distance $d_{\text{inf}} = \SI{0.02}{\meter}$.
    
    \item \textbf{Upright Posture} ($A_{\text{upright}}$):
    An equality constraint $A_{\text{upright}} \Delta q = 0$ ensures that the sum of velocities for \texttt{torso\_1}, \texttt{torso\_2}, and \texttt{torso\_3} remains zero for a vertical torso orientation throughout the motion. For the \textit{delivery} task, the velocities of these three joints are individually fixed to zero to maximize payload stability.
\end{itemize}

The specific weight parameters used during evaluation are summarized in Table~\ref{tab:wbik_params}. In our optimization hierarchy, maintaining the center-of-mass ($C_{\text{com}}$) over the base is given the highest priority to ensure global stability, followed by accurate bimanual end-effector tracking ($C_{\text{ee}}$). The relative priorities among the mobile base, torso, and arm configurations are tuned according to the specific requirements of each task to balance mobility, stability, and reachability.
During our experiments, these parameter settings required only 1–2 rollout checks before being finalized.

\begin{table}[t]
\centering
\caption{Whole-body IK parameters for the three evaluation tasks}
\label{tab:wbik_params}
\resizebox{\columnwidth}{!}{%
\begin{tabular}{lccc}
\toprule
Symbol & Laundry & Delivery & Tablescape \\
\midrule
$w_p$ & 10000 & 10000 & 10000 \\
$w_o$ & 10000 & 10000 & 10000 \\
$w_{\text{nom,torso}}$ & 50 & 1000 & 200 \\
$w_{\text{nom,arm}}$ & 50 & 1000 & 10 \\
$w_{\text{curr}}$ & 50 & 1000 & 10 \\
$w_{\text{base,pos}}$ & 50 & 50 & 5000 \\
$w_{\text{base,ori}}$ & 50 & 50 & 5000 \\
$w_{\text{com}}$ & 100000 & 100000 & 100000 \\
$b_x, b_y$ & \SI{0.08}{\meter} & \SI{0.08}{\meter} & \SI{0.08}{\meter} \\
\bottomrule
\end{tabular}%
}
\end{table}

\section{Robot Hardware Setup}
Our platform is built on a commercially available \texttt{Rainbow Robotics RB‑Y1} bimanual mobile manipulator, which provides a holonomic base with mecanum wheels, a 6‑DoF torso, two 7‑DoF arms, and a 2-DoF neck. The stock parallel grippers are replaced with fin‑ray fingers identical to the fingers of the UMI grippers.
On the 2‑DoF neck we mount a stereo pair of industrial wide‑angle RGB cameras (\texttt{FLIR BFS‑PGE‑23S3C‑CS}) using a custom neck bracket and pitch‑joint stop that constrains the neck range for safe cabling. On each wrist, we install a custom clamp and rigidly attach a \texttt{FLIR BFS‑PGE‑50S5C‑C} camera at a fixed offset relative to the gripper. 

\begin{figure}[t]
    \centering
    \includegraphics[width=\linewidth]{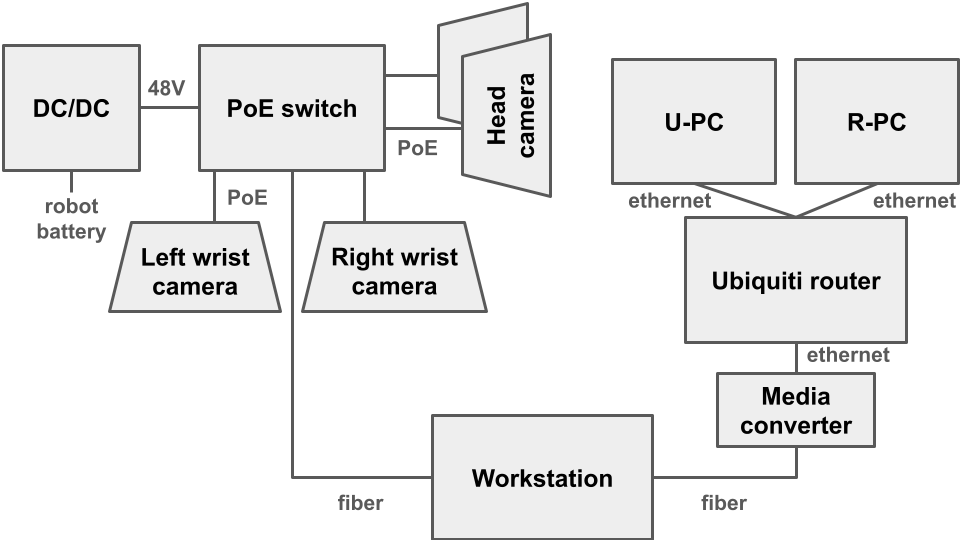}
    \caption{\textbf{Hardware Schematic.}}
    \label{fig:schematic}
\end{figure}

As shown in Fig.~\ref{fig:schematic}, our sensing and computing stack is organized around two fiber‑connected subnetworks. All cameras connect via GigE Ethernet cables to a PoE (Power over Ethernet) GigE switch mounted in a backpack structure on the robot torso. This PoE switch uplinks to the external workstation through a \SI{10}{Gbps} multimode fiber link, providing sufficient bandwidth for simultaneous multi‑camera streaming. In parallel, a Ubiquiti router connects the robot’s internal control computers -- including a user PC (U-PC) and a robot PC (R-PC) -- to the workstation over a second fiber link, carrying all low‑level robot state and command traffic. The external workstation runs our control stack, GigE camera drivers, and policy inference, and interfaces with the robot via these two fiber uplinks. This architecture supports high‑bandwidth multi-camera streaming and high-frequency robot closed‑loop control.

\section{Additional Experiments}
\subsection{Depth quality sensitivity}
\begin{figure}[h]
    \centering
    \vspace{2mm}
    \includegraphics[width=\linewidth]{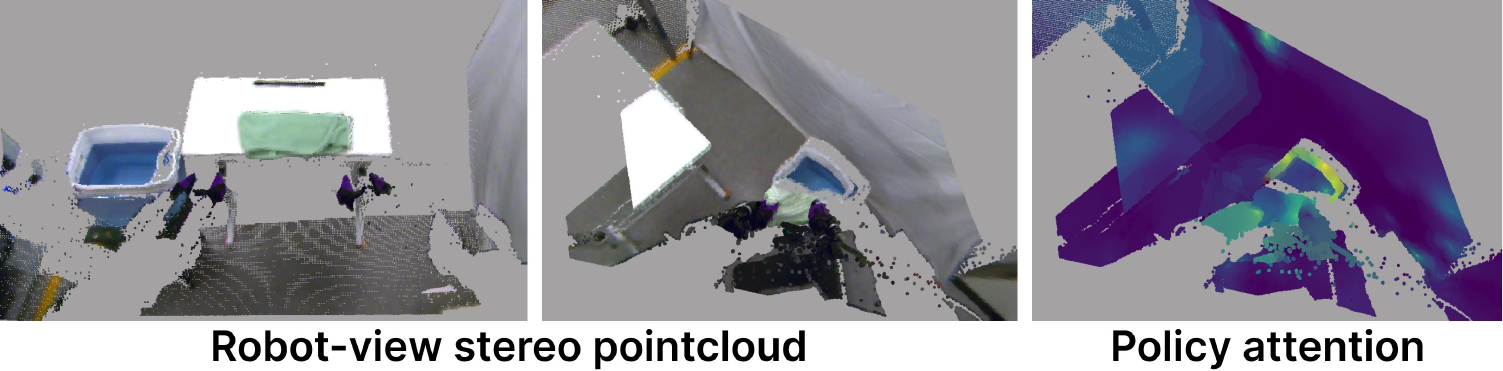}
    \caption{High quality stereo pointcloud and informative attention.}
    \label{fig:depth}
\end{figure}

Fig.~\ref{fig:depth} shows that the robot-view stereo pointcloud is clean and the policy attends to task-relevant regions (bin and cloth) during inference.

We inject additional Gaussian noise into depth and evaluate the Laundry task under 3 noise levels (10 rollouts each). 
As shown in Tab.~\ref{tab:result}, performance remains unchanged up to \SI{1}{cm} standard deviation noise and degrades only under substantially larger perturbations.
\begin{table}[h]
    \centering
    \setlength{\tabcolsep}{2pt}
    \vspace{2mm}
    \begin{tabular}{c|cccc}
        \hline
        noise std (mm) & 0 & 2 & 10 & 20 \\
        \hline
        success (\%) & 90 & 90 & 90 & 50 \\
        \hline
    \end{tabular}
    \caption{Depth quality sensitivity.}
    \label{tab:result}
\end{table}

\subsection{Robustness to unseen lightings and objects.}
To further assess robustness, we additionally evaluated the Laundry task under 3 unseen lighting conditions and on 3 unseen objects (4 rollouts per case). Object variation yields 
$90.63\%\pm12.10\%$ success, and lighting variation yields $93.75\%\pm10.83\%$, indicating robustness of our policy.

\begin{table}[H]
    \centering
    \vspace{2mm}
    \setlength{\tabcolsep}{2pt}
    \begin{tabular}{c|cccccccc}
        \hline
        \textbf{s}een and \textbf{u}nseen object & s1 & s2 & u1 & u2 & u3 & u4 & u5 & u6 \\
        \hline
        success (\%) & 100 & 75 & 100 & 75 & 100 & 100 & 100 & 75\\
        \hline
    \end{tabular}
    \label{tab:object}
    \caption{Generalization to unseen objects.}
\end{table}

\begin{table}[H]
    \centering
    \vspace{2mm}
    \setlength{\tabcolsep}{2pt}
    \begin{tabular}{c|cccc}
        \hline
        light (lux) & 4300 & 3370 & 1840 & 810 \\
        \hline
        success (\%) & 100 & 100 & 100 & 75 \\
        \hline
    \end{tabular}
    \caption{Robustness to lighting variations.}
    \label{tab:light}
\end{table}

\subsection{iPhone-ARKit tracking error.}
We measure the average tracking error against ground truth obtained from Motion Capture. The error is within 5.0 mm position / 0.8$\degree$ rotation.

\end{document}